\documentclass[]{article}

\usepackage[final]{neurips_2019}

\usepackage[utf8]{inputenc} 
\usepackage[T1]{fontenc}    
\usepackage{hyperref}       
\usepackage{url}            
\usepackage{booktabs}       
\usepackage{amsfonts}       
\usepackage{nicefrac}       
\usepackage{microtype}      

\usepackage[inline]{enumitem}
\usepackage{amsmath}
\usepackage{amssymb}

\usepackage{mathtools, stmaryrd}

\usepackage{hyperref, cleveref}
\hypersetup{colorlinks=true, citecolor=blue}

\setcitestyle{open={(},close={)}}

\usepackage{xparse} \DeclarePairedDelimiterX{\Iintv}[1]{\llbracket}{\rrbracket}{\iintvargs{#1}}
\NewDocumentCommand{\iintvargs}{>{\SplitArgument{1}{,}}m}
{\iintvargsaux#1} %
\NewDocumentCommand{\iintvargsaux}{mm} {#1\mkern1.5mu,\,\mkern1.5mu#2}

\newcommand{\indep}{\perp \!\!\! \perp}

\title{Simple Video Generation using Neural ODEs}
\author{%
  David Kanaa \thanks{Equal contribution. Names arranged alphabetically.} \\
  Polytechnique Montr\'eal, Mila \\
  \texttt{david.kanaa@gmail.com}
  \And
  Vikram Voleti  \footnotemark[1] \\
  Universit\'e de Montr\'eal, Mila \\
  \texttt{voletivi@mila.quebec} \\
  \And
  Samira Ebrahimi Kahou \\
  École de technologie supérieure, Mila, CIFAR \\
  \texttt{samira.ebrahimi-kahou@etsmtl.ca} \\
  \And
  Christopher Pal \\
  Polytechnique Montréal, Mila, CIFAR \\
  \texttt{christopher.pal@polymtl.ca}
}

\begin{document}

\maketitle

\begin{abstract}
    Despite having been studied to a great extent, the task of conditional generation of sequences of frames---or videos---remains extremely challenging. It is a common belief that a key step towards solving this task resides in modelling accurately both spatial and temporal information in video signals. A promising direction to do so has been to learn latent variable models that predict the future in latent space and project back to pixels, as suggested in recent literature. Following this line of work and building on top of a family of models introduced in prior work, Neural ODE, we investigate an approach that models time-continuous dynamics over a continuous latent space with a differential equation with respect to time.
    The intuition behind this approach is that
    these trajectories in latent space could then be extrapolated to generate video frames beyond the time steps for which the model is trained.
    We show that our approach yields promising results in the task of future frame prediction on the Moving MNIST dataset with 1 and 2 digits.
    

\end{abstract}

\section{Introduction}

Conditional frame generation in videos (interpolation and/or extrapolation) remains a challenging task despite having been well studied in the literature. It involves encoding the first few frames of a video into a good representation that could be used for subsequent tasks, i.e. prediction of the next few frames.
Solving the task of conditional frame generation in videos requires one to identify, extract and understand latent concepts in images, as well as adequately model both spatial and temporal factors of variation. Typically, an encoder-decoder architecture is used to first encode conditioning frames into latent space, and then recurrently predict future latent points and decode them into pixel space to render future frames.

In this paper, we investigate the use of Neural Ordinary Differential Equations~\citep{chen2018neural} (Neural ODEs) for video generation. The intuition behind this is that we would like to enforce the latent representations of video frames to follow continuous dynamics. Following a dynamic means that frames close to each other in the space-time domain (for example, any video of a natural scene) are close in the latent space. This implies that if we connect the latent embeddings of contiguous video frames, we should be able to obtain trajectories that can be solved for with the help of ordinary differential equations.

Since these trajectories follow certain dynamics in latent space, it should also be possible to extrapolate these trajectories in latent space to future time steps, and decode those latent points to predict future frames. In this paper, we explore this possibility by experimenting on a simple video dataset --- Moving MNIST~\citep{Srivastava2015UnsupervisedLO} --- and show that Neural ODEs do offer the advantages described above in predicting future video frames.

Our main contributions are:
\begin{itemize}
    \item we combine the typical encoder-decoder architecture for video generation with Neural ODEs,
    \item we show promising results on 1-digit and 2-digit Moving MNIST,
    \item we discuss the future directions of work and relevant challenges.
\end{itemize}

To the best of our knowledge, this is the first work that explores the use of Neural ODEs in the space of video generation.

\section{Related Work}
%
%
Early work on using deep learning algorithms to perform video generation has tackled the problematic in various ways, from using stacked regular LSTM layers~\citep{Srivastava2015UnsupervisedLO} to combining convolution with LSTM modules in order to extract local spatial information which correlates with long-term temporal dependencies~\citep{xingjian2015convolutional}.~\citep{prabhat2017deep} show that 3D convolution can be effectively used to extract spatio-temporal information from sequences of images for extreme weather detection.~\citep{wang2018video} use a generative model guided with segmentation maps to generate single step future frame. While their results may be interesting, due to the conditioning, the model might rely too much on segmentation.

Other work in the recent literature,~\citep{Babaeizadeh2017StochasticVV, denton18SVGLP, Lee2018StochasticAV} incorporate stochastic components to their model that encodes the conditioning frames into latent space similar to a prior distribution, which is then sampled from to predict the next frames. This is done in order to take into account the uncertainty over possible futures.

More recently, \citep{Castrejn2019ImprovedCV} show that using a hierarchy of latent variables to improve the expressiveness of their generative model can lead to noticeably better performance on the task of video generation. \citep{Clark2019EfficientVG} use a Generative Adversarial Network combined with separable spatial and temporal attention models applied on latent feature maps in order to  handle spatial and temporal consistency separately.

While some of the above methods have yielded state-of-the-art results, some still struggle to produce smooth motions and for those who do produce continuously smooth ones, they enforce it through temporal regularisation in the optimisation objective, or through a specific training procedure. Drawing from recent work on using parameterised ODE estimators~\citep{chen2018neural} to model continuous-time dynamics, we choose to approach this problem with the intuition that we would like the video frames to be smoothly connected in latent space according to some continuous dynamics which we would learn. Unlike recurrent neural networks and other purely auto-regressive approaches which require observations to occur at uniform intervals, and may require three different models to extrapolate forward and backwards, or interpolate, continuously-defined dynamics should naturally allow to process observations occurring at non-uniform intervals and generate at any time, thus reducing the number of models required to perform extrapolation and interpolation to one.

\section{Neural Ordinary Differential Equations}

Neural Ordinary Differential Equations~\citep{chen2018neural} (Neural ODEs) represent a family of parameterised algorithms designed to model the evolution across time of any system, of state $\boldsymbol{\xi}(t)$ at an arbitrary time $t$, governed by continuous-time dynamics satisfying a Cauchy (or initial value) problem
\begin{equation*}
     \begin{cases}
        \,\boldsymbol{\xi}(t_0) &= \boldsymbol{\xi}_0 \\ \\
        \,\dfrac{\partial\boldsymbol{\xi}}{\partial t}(t) &= f(\boldsymbol{\xi}(t), t)
     \end{cases}
\end{equation*}
By approximating the differential with an estimator $f_{\theta} \simeq f$ parameterized by $\theta$, such as a neural network, these methods allow to learn such dynamics (or, trajectories) from relevant data.
Thus formalised, the state $\xi(t)$ of such a system is defined at all times, and can be computed at any desired time using a numerical ODE solver, which will evaluate the dynamics $f_{\theta}$ to determine the solution.

\[ (\boldsymbol{\xi}_0, \boldsymbol{\xi}_1, \dots, \boldsymbol{\xi}_n) = \text{\ttfamily ODEsolver}(f_{\theta} , \boldsymbol{\xi}_0, (t_0, t_1, \dots, t_n)) \]

For any single arbitrary time value $t_i$, a call to the {\ttfamily ODEsolver} computes a numerical approximation of the integral of the dynamics from the initial time value $t_0$ to $t_i$.
\[ \boldsymbol{\xi}_i = \text{\ttfamily ODEsolver}(f_{\theta} , \boldsymbol{\xi}_0, (t_0, t_i)) \, \simeq \, \boldsymbol{\xi}_0 + \int_{t_0}^{t_i} f_{\theta}\left(\boldsymbol{\xi}(s), s\right)\,ds \, = \, \boldsymbol{\xi}(t_i)\]

There exist in the literature a plethora of algorithms to perform numerical integration of differential equations. Amongst the most common are : the simplest, Euler's method; higher order methods such as Runge-Kutta methods~\citep{pontyagin1962mathematical, kutta1901beitrag,hairer2000solving}; as well as multistep algorithms like the Adams-Moulton-Bashforth methods~\citep{butcher2016numerical, hairer2000solving, quarteroni2000matematica}. More advanced algorithms have been proposed to provide better control over the approximation error and the accuracy~\citep{press2007numerical}.
In their implementation\footnote{https://github.com/rtqichen/torchdiffeq}~\citep{chen2018neural} use a variant of the fifth-order Runge-Kutta with adaptive stepsize and monitoring of the local truncation error to ensure accuracy.

The optimisation of the Neural ODE is performed through the framework of adjoint sensitivity~\citep{pontyagin1962mathematical} which can be formalised as follows : provided a scalar-valued objective function
\[L(\theta) = L\left(\boldsymbol{\xi}_0 + \int_{t_0}^{t_i} f_{\theta}\left(\boldsymbol{\xi}(s), s\right)\,ds\right)\]
the gradient of the objective with respect to the model's parameters follows the differential system
\begin{align*}
        \dfrac{d\boldsymbol{a}(t)}{dt}   &= - \,\, \boldsymbol{a}(t)^\top \, \dfrac{\partial f(\boldsymbol{\xi}(t), t, \theta)}{\partial \boldsymbol{\xi}} \\
        \dfrac{dL}{d\theta} \,\,&= - \int_{t_i}^{t_0} \boldsymbol{a}(s)^\top \, \dfrac{\partial f(\boldsymbol{\xi}(s), s, \theta)}{\partial \theta} ds
\end{align*}
where the $\boldsymbol{a}(t) = \partial L/\partial \boldsymbol{\xi}$ is the adjoint.

\section{Approach}

Our approach combines the familiar encoder-decoder architecture of neural network models with a Neural ODE that works in the latent space.
\begin{enumerate}
    \item We encode the conditioning frames into a point in latent space
    \item We feed this latent embedding to a Neural ODE as the ``initial value'' at time $t=0$, and use it to predict latent points corresponding to future time steps.
    \item We decode each of these latent points into frames in pixel space at different time steps
\end{enumerate}

More formally, in accordance with established formulations of the task of video prediction, let us assume a setting in which we have a set of $m$ contextual frames $\mathcal{C} = \{ (x_{i}, t_{i}) \}_{i \in \Iintv{0, m}}$. We seek to learn a predictive model such that, provided $\mathcal{C}$, we can make predictions $\mathcal{P}(\mathcal{C})= \{ (x_{j}, t_{j}) \}_{j \in \Iintv{m, m+n}}$ about the evolution of the video across time, arbitrarily in the future or past (extrapolation) or even in between observed frames (interpolation).

Let $x(t)$ denote the continuous signal representing the video stream from which $\mathcal{C}$ is sampled, that is : \[\forall (x_i, t_i) \in \mathcal{C}, \, x(t_i) = x_i\]
The temporal changes in the raw signal $x(t)$ can be interpreted as effects of temporal variations in the latent concepts embedded within it. For example, suppose we have a video of a ball moving, any temporal change in the video will be observed only on pixels related to the latent notion of "moving ball". Because the concept "ball" follows some motion, the related pixels will change accordingly.
From this statement it follows the intuition to model dynamics in latent space and capture spatial characteristics separately. Thus we learn a predictor $\mathcal{P}$ which 
\begin{enumerate*}[label=$(\roman*)$]
      \item learns a latent representation of the observed discrete sequence of frames that captures spatial factors of variation, as well as
      \item infers plausible latent continuous dynamics from which the aforementioned discrete sequence may be sampled i.e. which better explains the temporal variations within the sequence.
\end{enumerate*}

The proposed model follows the formalism of latent variable model proposed by~\citep{chen2018neural} in which the latent at the current time value $z(t_{m})$ is sampled from a distribution $\mathbb{P}_{Z}$, the latent generative process is defined by an ODE that determines the trajectory followed in latent space from the initial condition $z(t_{m})$, and a conditional $\mathbb{P}_{X|Z}$ with respect the latent vectors predicted along the trajectory at provided times is used to independently sample predicted images:

\vspace{-0.5em}

\begin{align*}
        z_{m} &\sim \mathbb{P}_{Z} \left( \cdot \right) \\
        z(t_{i}) &= \mathcal{I} (f_{\theta}, z_{m}, t_{m}, t_{i}) = z_{m} + \int_{t_{m}}^{t_{i}} f\left(z(s), s; \theta \right)\,ds &\forall t_{i} &\in \Iintv{t_{m}, t_{m+n}} \\
        x(t_{i}) &\sim \mathbb{P}_{X|Z} \left( \cdot\,|\, z(t_{i}) \right),\quad x(t_{i}) \indep x(t_{j}) &\forall t_{i}, t_{j} &\in \Iintv{t_{m}, t_{m+n}}
\end{align*}

In practice, we use an approximate posterior $q_{\phi}(\cdot\,|\,\mathcal{C})$ instead of $\mathbb{P}_{Z}$, and similarly, instead of $\mathbb{P}_{X|Z}$, we use an estimator $p_{\psi}(\cdot\,|\,z(t_{m}))$. Together, these estimators function as an \emph{encoder-decoder} pair between the space of image pixels and that of latent representations.

We investigate a deterministic setting where a unique and non-recurrent pair encoder-decoder is used to process every frame. The encoder projects a frame $(x_{i}, t_{i})$ onto an embedding $z_{t_{i}} = z_{i} = q_{\phi}(x_{t_{i}})$, then the ODE defining the latent dynamics is integrated to produce the value of the latent embedding $z_{t_{i}} = \mathcal{I} (f_{\theta}, z_{0}, t_{0}, t_{i})$. Finally, the decoder is used to project $z(t_{j})$ back into an image $\hat{x}_{t_{j}} = p_{\psi}(z_{t_{j}})$. In terms of objective function used to optimise the parameters of the model, we use a combination of an $L_2$ reconstruction in pixel space, and an $L_2$ distance between the latent points predicted by the NeuralODE and the embeddings of each frame:

\vspace{-0.5em}

\begin{multline}\label{eq:loss2}
    \mathcal{L}(\phi, \theta, \psi) = \,\sum_{\Iintv{t_{m}, t_{m+n}}} \| x_{t_{i}} - p_{\psi} \circ \mathcal{I} (f_{\theta}, q_{\phi}(x_{t_{0}}), t_{0}, t) \|_{2}^{2} \\
    + \| q_{\phi}(x_{t_{i}}) - \mathcal{I} (f_{\theta}, q_{\phi}(x_{t_{0}}), t_{0}, t_{i}) \|_{2}^{2}
\end{multline}

The latter component of the objective function is meant to ensure that we learn a compact latent subspace to which both the learnt dynamics and the encoder project. More precisely, it enforces the latent representation predicted by the Neural ODE to match that estimated for each time step by the encoder.

We also inquire into the sequence-to-sequence architecture~\citep{chen2018neural}, where

\begin{align*}
    \mathbb{P}_{Z} = \mathcal{N}(\mu(\mathcal{C}), \sigma^{2}(\mathcal{C})), &\quad \mathbb{P}_{X|Z} = \mathcal{N}(\mu(z(t_{m})), \sigma^{2}(z(t_{m}))) \\
    &\text{thus,} \\
    q_{\phi}(\cdot\,|\,\mathcal{C}) = (\mu_{\phi}(\mathcal{C}),\sigma^{2}_{\phi}(\mathcal{C})), &\quad p_{\psi}(\cdot\,|\,z(t_{m})) = (\mu_{\psi}(z(t_{m})), \sigma^{2}_{\psi}(z(t_{m})))
\end{align*}

In practice, $\sigma_{\psi}(z(t_{m}))$ is set to a constant value $\sigma = 1$ and $\mu_{\psi}(z(t_{m})) = x_{m}$, the true frame observed at time $t_{m}$. In this setting, the variational encoder $q_{\phi}$ used is based on an RNN model over the context $\mathcal{C} = \{ (x_{i}, t_{i}) \}_{i \in \Iintv{0, m}}$, whereas the decoder $p_{\psi}$ is non-recurrent---hypothesis of independence between generated frames; the temporal dependencies are modelled by the ODE---. At training time, the entire estimator is optimised as a variational auto-encoders~\citep{kingma2013auto, rezende2014stochastic} through the maximisation of the Evidence Lower Bound (ELBO):

\begin{equation}\label{eq:elbo}
    \mathcal{E}(\phi, \theta, \psi) = \,\underbrace{\sum_{\Iintv{t_{m}, t_{m+n}}} -\,\mathbb{E}_{z_{m} \sim q_{\phi}(\cdot\,|\,\mathcal{C})} [\log p_{\psi}(\hat{x}_{t}\,|\,\mathcal{I} (f_{\theta}, z_{m}, t_{m}, t)]}_{\text{reconstruction term}}
    + \mathrm{D}_{KL}(q_{\phi}(\cdot\,|\,\mathcal{C})\,\|\,\mathcal{N}(0, \boldsymbol{I}))
\end{equation}

\section{Experiments on Moving MNIST}

We explore two different methods of combining an encoder-decoder framework with ODEs for 1-digit and 2-digit Moving MNIST~\citep{Srivastava2015UnsupervisedLO}. In each case, we use the first 10 frames as both input to the model and as ground truth for reconstruction, which is the output of the model. We then check how the model performs on the subsequent 10 frames.

\subsection{1-digit Moving MNIST with non-RNN Encoder}

\begin{figure*}[t]
\centering
\begin{tabular}{cc}
\includegraphics[width=.45\textwidth]{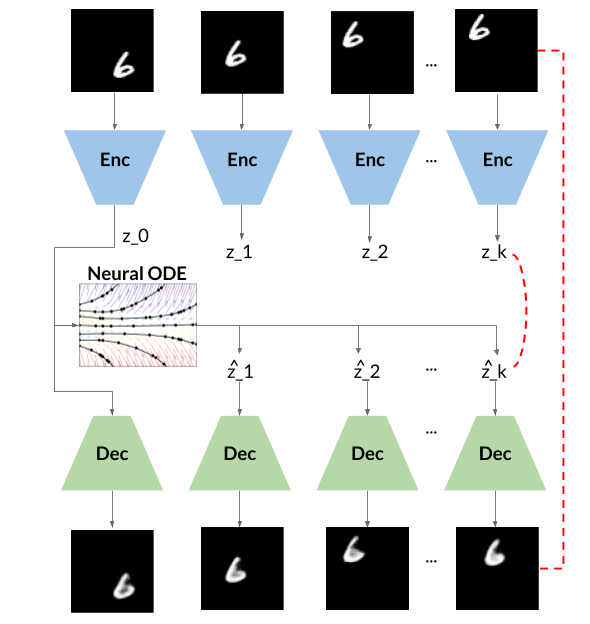} &
\includegraphics[width=.45\textwidth]{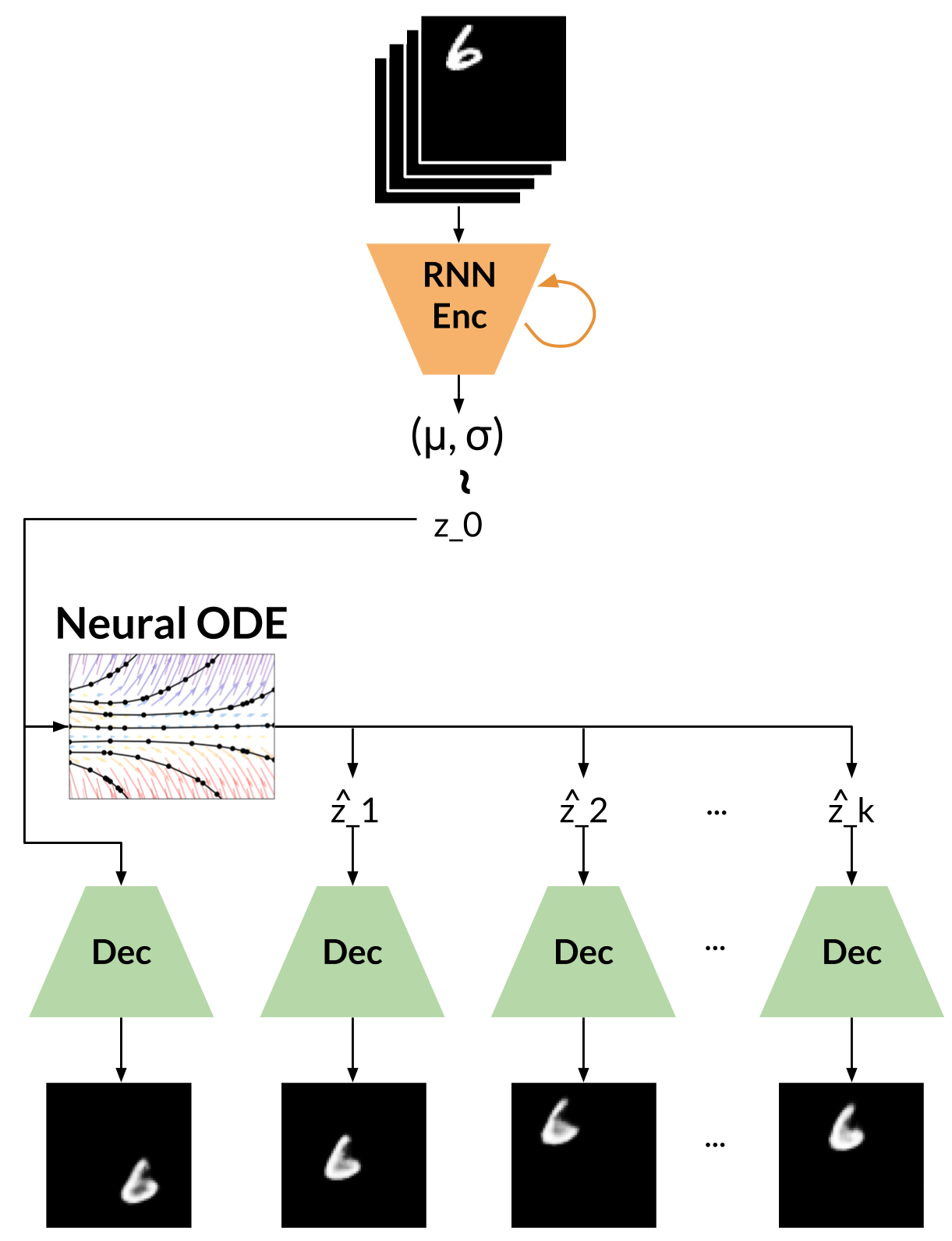}\\
(a) & (b)
\end{tabular}
\caption{Architectures for Encoder-ODE-Decoder}
\label{fig:arch}
\end{figure*}

This method, corresponding to \autoref{eq:loss2}, involves an encoder and a decoder that each act on a single frame to embed and decode, respectively, a latent representation. Figure~\ref{fig:arch} (a) shows this architecture. Here, we try to enforce this representation to follow a continuous dynamics in latent space such that there is a one-to-one mapping between the raw pixel space and the latent space from both the encoder side as well as the decoder side.

This model takes one frame as the conditioning input, encodes it, feeds it to the ODE which then predicts the latent representations of the first 10 time steps (including the one which was fed to it), each of which is then decoded to pixel space. We then compute a loss between the reconstructed output and the original input. In addition, each frame of the original video is also encoded separately, and we compute another loss on the encoded latent representations and those predicted by the ODE. This is to enforce the latent representations provided by the encoder to follow the dynamics implicit in the Neural ODE.

\begin{figure*}[t]
\centering
\begin{tabular}{cc}
\includegraphics[width=.95\textwidth]{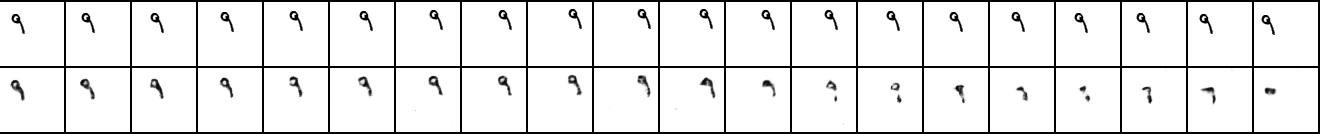}\\
(a) Train \\
\includegraphics[width=.95\textwidth]{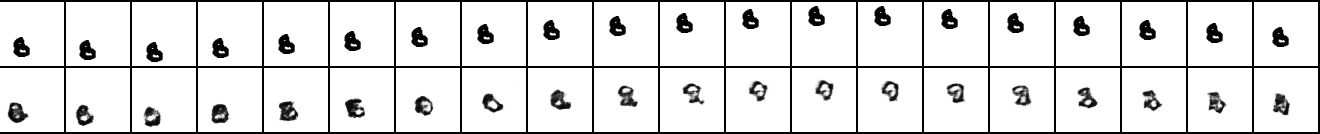}\\
(b) Validation
\end{tabular}
\caption{Samples predicted at 20 time steps, conditioned on the first 10 time steps with frames from (a) train set and (b) validation set using Non-RNN Encoder --- ODE --- Decoder (Figure~\ref{fig:arch}a). In each, top row are original samples, and bottom row are predicted samples. For this figure, we use the trained model to reconstruct the first 10 frames and then predict the next 10 frames}
\label{fig:arch1_samples}
\end{figure*}

We used 1000 video sequences of length 10 as conditioning input (as well as reconstruction output), and a batch size of 100. The encoder and decoder have inverted architectures with the same number of channels in their respective orders. Figure~\ref{fig:arch1_samples} shows samples from using this architecture.

\subsection{1-digit Moving MNIST with RNN Encoder}

While the previous architecture works pretty well on the training samples, We see that it does not work very well on validation data. We believe that there are two things that must be corrected:

\begin{itemize}
    \item The encoder must be conditioned on multiple frames.
    \item The latent representation provided by the encoder must be stochastic in nature
\end{itemize}

Since the Neural ODE is only seeing the first frame of the video to base its latent dynamic trajectories on, it is a highly constrained problem.
However, we would like to relax this constraint by conditioning the Neural ODE on multiple frames, which is commonly practiced in video prediction/generation.

We would also like to make the model stochastic. The previous model is deterministic, so there is a high chance it simply memorizes the training data. So, given an input frame, there is exactly 1 trajectory the Neural ODE is able to generate for it, so there is no scope of any variation in the generated videos. We would like to generate different videos given the same conditioning input, since it matches with real world data.

Figure~\ref{fig:arch} (b) shows such an architecture that solves both the above issues, corresponding to \autoref{eq:elbo}. It is similar to a Variational Recurrent Neural Network~\citep{Chung2015ARL}, except here a Neural ODE handles the latent space.

We feed the first 10 frames as conditioning input to a Recurrent Neural Network (RNN). This network outputs the mean and variance of a multivariate Gaussian distribution. We sample a latent point from this distribution, and feed this to a Neural ODE as the initial value of the latent variable at $t=0$. The Neural ODE then predicts latent representations at the first 10 time steps, which we then decode independently to raw pixels. We compute a reconstruction loss between the predicted frames and the original frames in the first 10 time steps. We also add a KL-divergence loss between the predicted Gaussian distribution and the standard normal distribution, to constrain the latent representation to follow a standard normal prior.

\begin{figure*}[t]
\centering
\begin{tabular}{cc}
\includegraphics[width=.95\textwidth]{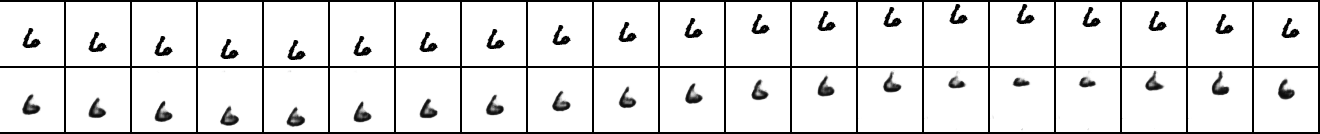}\\
\includegraphics[width=.95\textwidth]{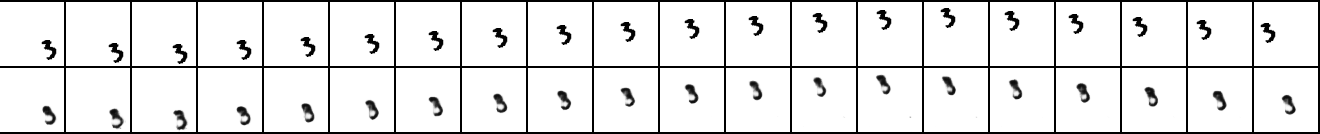}\\
(a) Train \\
\includegraphics[width=.95\textwidth]{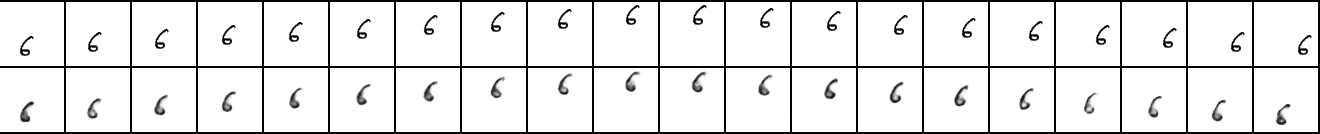}\\
\includegraphics[width=.95\textwidth]{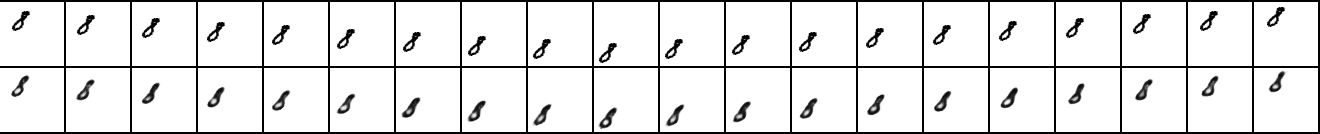}\\
\includegraphics[width=.95\textwidth]{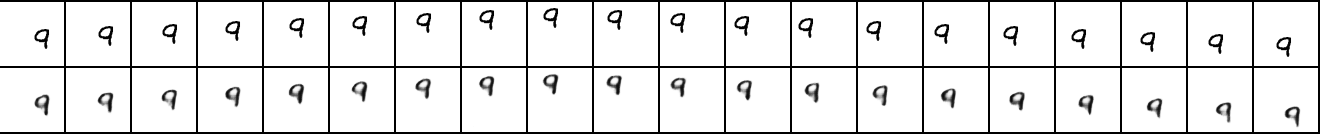}\\
\includegraphics[width=.95\textwidth]{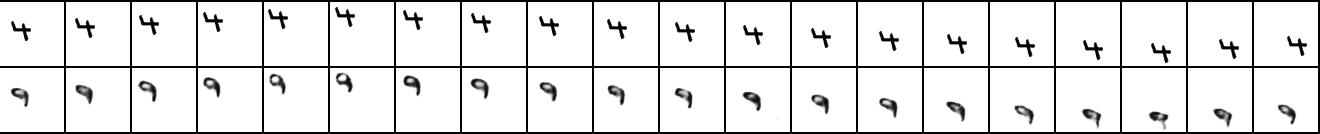}\\
\includegraphics[width=.95\textwidth]{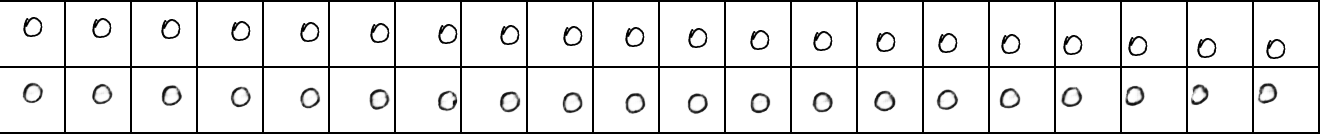}\\
(b) Validation
\end{tabular}
\caption{Samples predicted at 20 time steps, conditioned on the first 10 time steps with frames from (a) train set and (b) validation set using RNN Encoder --- ODE --- Decoder (Figure~\ref{fig:arch}b). In each, top row are original samples, and bottom row are predicted samples. For this figure, we use the trained model to reconstruct the first 10 frames and then predict the next 10 frames}
\label{fig:arch2_samples}
\end{figure*}

The model architectures of the encoder (except the recurrent part) and the decoder are the same as in the previous model. We provide 10000 videos as training input, and use a batch size of 128. Figure~\ref{fig:arch2_samples} shows the results using this architecture. We can see that the model has been able to capture both structural information and temporal information.


\subsection{2-digit Moving MNIST with RNN Encoder}

We use the same architecture as for 1-digit Moving MNIST (Figure~\ref{fig:arch} (b)) to try to reconstruct 2-digit Moving MNIST. We used the same model settings (number of layers, number of channels, etc.) and the same optimization settings. At the time of writing this paper, we stopped the training at 2000 epochs, same as that for 1-digit Moving MNIST.

\begin{figure*}[t]
\centering
\begin{tabular}{cc}
\includegraphics[width=.95\textwidth]{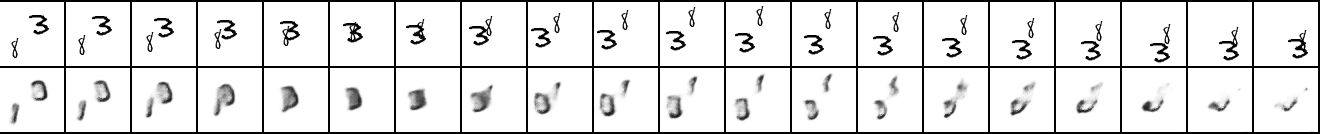}\\
(a) Train \\
\includegraphics[width=.95\textwidth]{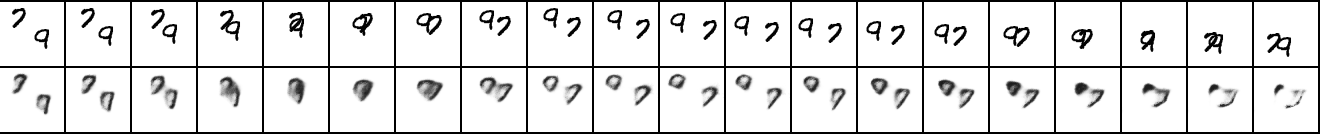}\\
\includegraphics[width=.95\textwidth]{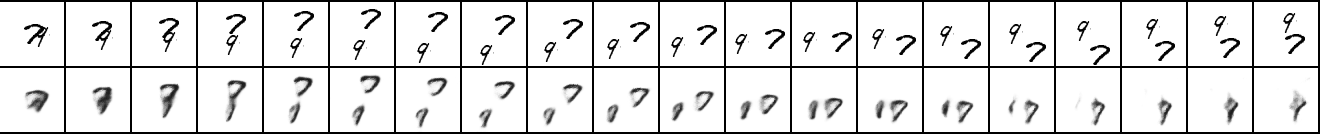}\\
(b) Validation
\end{tabular}
\caption{(top) Original and (bottom) predicted samples of RNN Encoder --- ODE --- Decoder on 2-digit Moving MNIST}
\label{fig:mmnist2_samples}
\end{figure*}

Figure~\ref{fig:mmnist2_samples} shows some samples from a model trained on 2-digit Moving MNIST. We believe the spatial trajectories of each individual digit are being recorded very well by the Neural ODE. However it would take many more epochs for the encoder and decoder to reconstruct the images better. This phenomenon of the Neural ODE training earlier than the encoder-decoder was observed while training 1-digit Moving MNIST as well.

\subsection{A note on the problem formulation}

It is to be noted that there is a difference between our problem formulation for training our model, and the usual one. The typical way of generating videos is to condition a model on the first few frames, and train it to predict some time steps in the future. Then in evaluation, the model is conditioned on the first few frames of unseen videos, and used to predict for time steps including and beyond those for which it was trained on.

In our training procedure, we condition our model on the first few frames, and then simply reconstruct those same frames. Hence, we formulate the training problem as \textit{reconstruction} instead of prediction. Despite this, at evaluation, we are able to predict future frames very well. This is because the dynamics followed in the set of frames used at training are preserved throughout the subsequent time steps.
And so by following the trajectory in latent space and decoding it, we are able to predict future frames.  

\section{Future work}

There are several future directions we are looking at:
\begin{itemize}
    \item We would first like to improve the results for 2-digit Moving MNIST. As of the time of writing this paper, we are already making progress in this direction.
    
    \item \textbf{Scaling up}: We would like to scale this up to bigger datasets such as KTH~\citep{kth}, the Kinetics dataset~\citep{kay2017kinetics}, etc. As of the time of writing this paper, we are 
    already making progress in this direction.
    
    \item \textbf{Fair comparison}: We would like to explore how well it performs when conditioning on some frames and training to predict the subsequent frames, as it is typically tackled by many of the recent papers on video generation~\citep{denton18SVGLP, Babaeizadeh2017StochasticVV, Lee2018StochasticAV}, so we can make a fair comparison of our approach with these methods.
    
    \item \textbf{Disentanglement}: We would like to examine the latent representation created by the Neural ODE in this domain in more detail. We would like to explore whether it implicitly disentangles spatial and temporal information, which it seems to be doing so from the evidence so far.
    
    \item \textbf{Visualization}: We would like to visualize the latent representation in lower dimensional space, to check the evidence of trajectories as being enforced by the Neural ODE. The best reason to use Neural ODE in the pipeline is so that the latent representation is now more interpretable --- consecutive time steps lie on a lower-dimensional trajectory. We would like to prove that this exists, and show how exploring in latent space maps to intuitive changes in the decoded frames. We plan on using tools such as t-SNE~\citep{vanDerMaaten2008} and UMAP~\citep{McInnes2018UMAPUM} for the visualization.
    
    \item \textbf{Temporal interpolation of videos}: Since videos follow continuous dynamics in latent space, it is possible to sample latent points for fractional time steps, i.e. time steps that are in the continuous range \textit{between} the time steps of the original video, and decode them using the same decoder. Hence, it should be possible to increase the frame rate of any given video without the any additional effort. This also opens the door to exploration of the capacity of the learned representation for other downstream tasks in video.
    
    \item \textbf{Better metric for evaluation}: We would also like to have a better metric to estimate how good the generated videos are. As mentioned earlier and as talked about in other papers~\citep{Lee2018StochasticAV}, the shortcomings of the current metrics of PSNR and SSIM are that they do not account for variation in the generated video from the ground truth. Since many recent models have a stochastic component, it is all the more important to be able to indicate that the generated video is different from the ground truth, but matches the data distribution of the ground truth. More recently, \citep{Clark2019EfficientVG} use the popular image quality metrics of Inception Score~\citep{salimans2016improved} and FID~\citep{Heusel2017GANsTB}, however, these metrics do not necessarily account for consistency in temporal information.
\end{itemize}

\section{Conclusion}

In this paper, we explored the use of Neural ODEs for video generation. We showed very promising results on the 1-digit and 2-digit Moving MNIST dataset. We investigated two different architectures, with and without a recurrent component, respectively stochastic and deterministic. Even though we formulated the training problem as reconstruction, we were able to use our model for prediction of future frames because we learn the continuous-time dynamics governing the temporal evolution of the latent features, using Neural ODEs.
We discussed in detail many future directions that would be useful to support our current approach, as well as help the space of video generation. We also discussed how our approach could be directly applied to other tasks such as temporal interpolation of videos. We hope that the research community uses our approach and takes advantage of the implicit feature of Neural ODEs to model continuous dynamics. We plan to release the code for all our experiments.

\bibliography{references}

\end{document}